\documentclass[letterpaper, 10 pt, conference]{ieeeconf}  

\IEEEoverridecommandlockouts                              

\usepackage{graphicx} 
\usepackage{epsfig} 
\usepackage{amsmath} 
\usepackage{algorithm}
\usepackage{algpseudocode}
\usepackage{caption}
\usepackage{subcaption}
\usepackage{amssymb}
\usepackage{booktabs}
\usepackage{multirow}
\usepackage{balance} 
\graphicspath{{images/}}

\title{
\Large \bf
GauTOAO: Gaussian-based Task-Oriented Affordance of Objects
}

\author{Jiawen Wang, Dingsheng Luo, \textit{Member, IEEE}\\
School of Intelligence Science and Technology, Peking University
}

\begin{document}

\maketitle
\thispagestyle{empty}
\pagestyle{empty}

\begin{abstract}
When your robot grasps an object using dexterous hands or grippers, it should understand the Task-Oriented Affordances of the Object(TOAO), as different tasks often require attention to specific parts of the object. To address this challenge, we propose GauTOAO, a Gaussian-based framework for Task-Oriented Affordance of Objects, which leverages vision-language models in a zero-shot manner to predict affordance-relevant regions of an object, given a natural language query. Our approach introduces a new paradigm: "static camera, moving object," allowing the robot to better observe and understand the object in hand during manipulation. GauTOAO addresses the limitations of existing methods, which often lack effective spatial grouping, by extracting a comprehensive 3D object mask using DINO features. This mask is then used to conditionally query gaussians, producing a refined semantic distribution over the object for the specified task. This approach results in more accurate TOAO extraction, enhancing the robot’s understanding of the object and improving task performance. We validate the effectiveness of GauTOAO through real-world experiments, demonstrating its capability to generalize across various tasks.

\end{abstract}

\section{INTRODUCTION}

Robot manipulation is a critical and well-established task in the field of robotics, integrating various aspects such as perception, control, and planning. While significant progress has been made in task-oriented grasping(TOG)\cite{Patankar_Phi_Mahalingam_Chakraborty_Ramakrishnan,Murali_Liu_Marino_Chernova_Gupta_2020}, where the focus is on the grasping affordance of objects for specific tasks, these studies have primarily concentrated on how to grasp an object rather than on the subsequent interactions required to complete the task. The broader challenge of understanding how to manipulate different parts of an object based on task requirements has received less attention in the literature.

\begin{figure}[t]
    \centering
    \includegraphics[width = 0.45\textwidth]{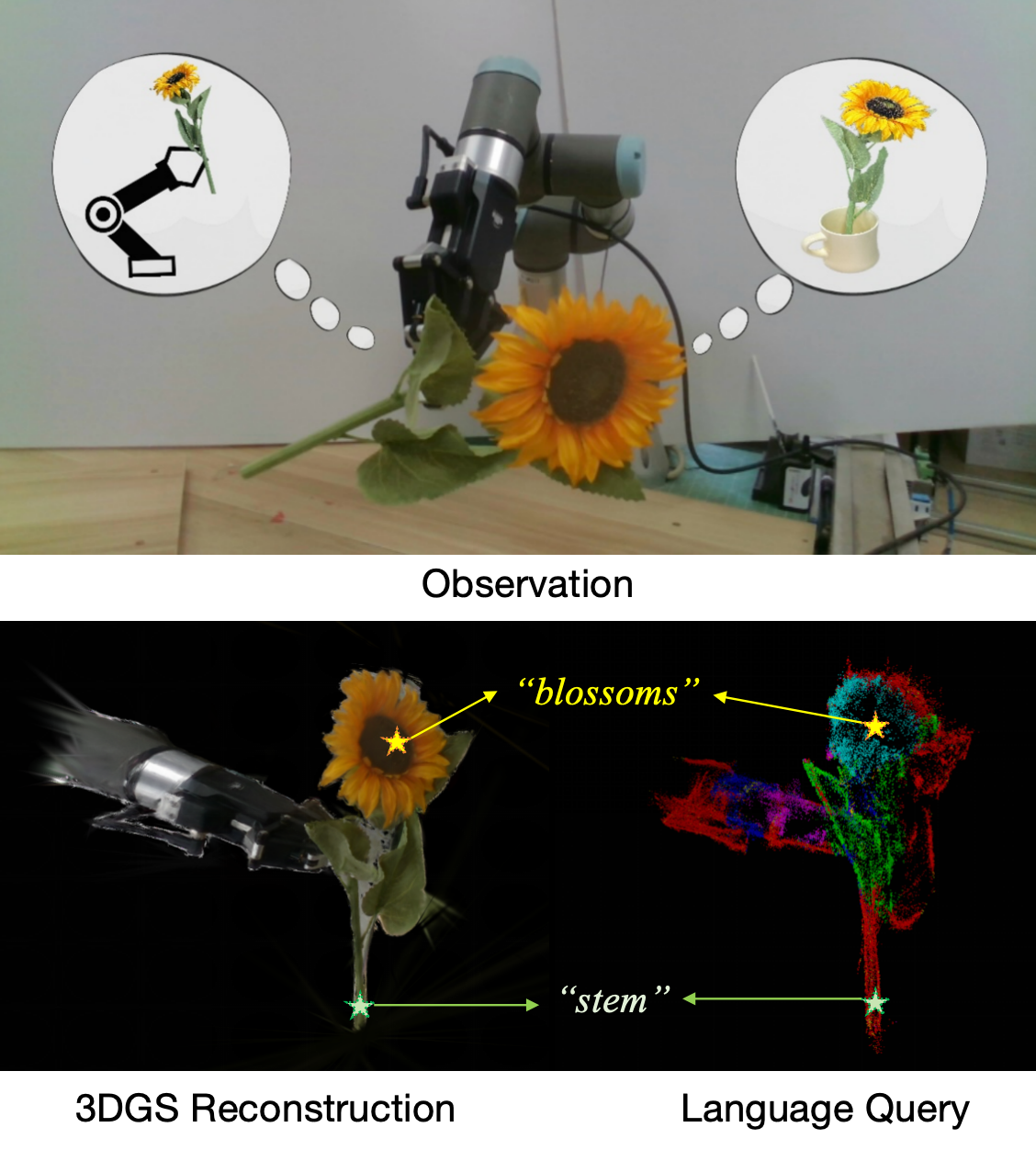}
    \caption{\textbf{GauTOAO.} When the robot is presented with a bouquet of sunflowers, its attention shifts based on the intended task. If the task is to display the flowers, the robot should focus on orienting the blossoms toward the observer. Conversely, if the task is to insert the flowers into a vase, the robot's focus shifts to aligning the flower stem with the vase's opening.}
    \label{fig:top}
\end{figure}

Task-oriented object manipulation introduces additional complexity compared to conventional grasping tasks. After an object is grasped, the robot needs to dynamically adjust its focus depending on the specific task at hand. As shown in Figure \ref{fig:top}, when the robot is presented with a bouquet of sunflowers, its attention shifts based on the intended task. For example, if the task is to display the flowers for appreciation, the robot should focus on orienting the blossoms toward the observer. Conversely, if the task is to insert the flowers into a vase, the robot's focus shifts to aligning the flower stem with the vase's opening. This ability to adaptively focus on different parts of the object in response to task demands highlights what we refer to as the \textbf{Task-Oriented Affordance of Objects(TOAO)}.

This concept extends beyond traditional grasping affordance, requiring robots to not only identify the optimal grasp but also to adapt their manipulation strategy based on the object’s different parts and their relevance to the task. By introducing this broader scope of task-oriented affordance, we address a gap in existing research and propose a framework that integrates dynamic perception and control mechanisms to facilitate object manipulation in diverse task contexts.

Perception is a critical upstream task in robot manipulation. Beyond traditional methods with explicit models, recent years have seen a rise in learning-based implicit models for robotic tasks, including techniques such as NeRF (Neural Radiance Fields)\cite{Mildenhall_Srinivasan_Tancik_Barron_Ramamoorthi_Ng_2020} and 3D Gaussian Splatting (3DGS)\cite{17Kerbl}. Building upon these, methods like LERF (Language-Embedded Radiance Fields)\cite{Kerr_Kim_Goldberg_Kanazawa_Tancik_Berkeley} and LangSplat\cite{qin2024langsplat} have emerged, embedding semantic information to achieve impressive advancements in perception. However, these approaches predominantly operate under the assumption of a “static object, moving camera” setup. This significantly diverges from the current trajectory of robotic development; for instance, it is unrealistic to expect a humanoid robot to walk around an object to study it. Drawing inspiration from human motor cognition, we propose a new paradigm: “static camera, moving object.” In this framework, a robot can manipulate and rotate the object with its arm to observe and perceive the object’s task-oriented affordance.

In this work, we introduce \textbf{GauTOAO} (\textbf{Gau}ssian-based \textbf{T}ask-\textbf{O}riented \textbf{A}ffordance of \textbf{O}bjects), a method that enables robots to understand and utilize the task-oriented affordances of objects based on natural language queries.

\begin{figure*}[!t]
    \centering
    \includegraphics[width=0.9\linewidth]{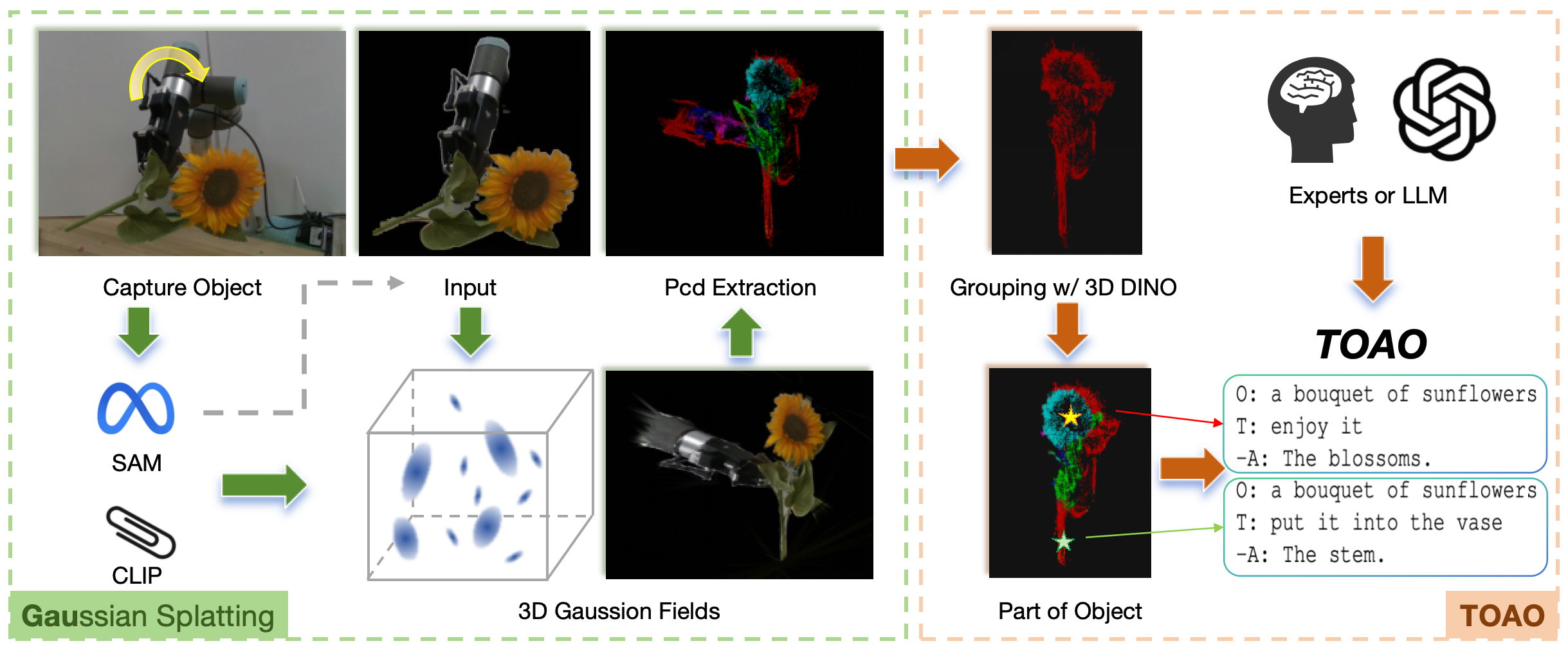}
    \caption{\textbf{The framework of GauTOAO.} The framework comprises two main components: 3D Gaussian splatting for reconstruction and TOAO extraction. In the first component, our approach builds upon LangSplat\cite{qin2024langsplat}, with a novel paradigm of a "static camera, moving object" setup. This paradigm involves the integration of SAM and depth information for image preprocessing, ensuring precise object reconstruction during manipulation. After training the model, the second stage focuses on TOAO extraction. By employing DINO's grouping first, we achieve more accurate extraction of the task-oriented affordance of objects.}
    \label{fig:framework}
\end{figure*}

GauTOAO takes as input both an in-hand object and a task in natural language (e.g., "O: flower; T: put it into the vase") and outputs the specific affordance-relevant part of the object (e.g., "stem" in this case), as illustrated in Figure \ref{fig:framework}. Our approach builds on the recent advances of 3D Language Gaussian Splatting (LangSplat) \cite{qin2024langsplat}, which utilizes calibrated RGB images and a collection of 3D Gaussians—each encoding language features distilled from CLIP \cite{radford2021learning}—to represent the language field. Given a query, LangSplat generates a 3D point cloud that highlights regions of the object based on their similarity to the query. However, LangSplat may sometimes highlight only partial object regions (e.g., focusing solely on the stem of the flower), which can be problematic for task-oriented scenarios where a comprehensive understanding of the object is essential. GauTOAO addresses this limitation by enhancing the 3D object understanding during inference. Specifically, it predicts a complete 3D object mask by leveraging 3D features extracted from DINO \cite{Caron_Touvron_Misra_Jegou_Mairal_Bojanowski_Joulin_2021}. This improves the robot’s capability to fully comprehend the in-hand object, enabling more accurate TOAO extraction.

\section{RELATED WORKS}

\subsection{Task-Oriented Affordance}
Early work on affordance learning has primarily focused on identifying the possible actions an object affords based on its shape and physical properties\cite{montesano2008learning}. Studies such as Gibson's affordance theory\cite{gibson1977theory} laid the foundation for understanding how humans and robots can perceive object affordances. Building on this, robotic research has explored how robots can learn affordances through both supervised learning and self-supervised exploration. For instance, Ashutosh et al.\cite{koppula2013learning} developed a framework where robots learn affordances by interacting with objects, using the observed outcomes to predict future interactions. Later works\cite{dang2014semantic}, utilized deep learning techniques to infer affordances directly from visual data, enabling robots to recognize and manipulate objects in unstructured environments. These works do not account for task-specific needs that may require a robot to focus on different parts of the same object for different tasks, a gap that Task-Oriented Affordance seeks to address.

\subsection{3D Gaussian Splatting}
Real-time neural rendering has been a long-standing goal, particularly for achieving high-quality visuals at interactive speeds. Kerbl et al.\cite{17Kerbl} proposed 3D Gaussian Splatting, enabling real-time 1080p rendering with state-of-the-art quality. This method has since been extended to various tasks, leveraging its efficient rendering process.

For dynamic scenes, works like Luiten et al.\cite{26Luiten_Kopanas_Leibe_Ramanan} introduced Dynamic 3D Gaussians, modeling changes over time, while Yang et al.\cite{43Yang_Gao_Zhou_Jiao_Zhang_Jin_2023} developed deformable 3D Gaussians, learning from a canonical space and modeling dynamic deformations. These adaptations expanded the technique to handle dynamic environments efficiently.

Moreover, 3D Gaussian Splatting has been integrated with diffusion models for text-to-3D generation\cite{10Chen_Wang_Liu,38Tang_Ren_Zhou_Liu_Zeng_2023,46Yi_Fang_Wu_Xie_Zhang_Liu_Tian_Wang}, such as DreamGaussian\cite{38Tang_Ren_Zhou_Liu_Zeng_2023}, which enhances 3D content creation.

\subsection{3D Open Vocabulary Understanding}

Recent advancements in 3D Gaussian Splatting have expanded its use beyond rendering to tasks such as segmentation and feature extraction. LEGaussians\cite{15Shi_Wang_Duan_Guan_2023} introduces uncertainty and semantic attributes to each Gaussian, enabling semantic map rendering with corresponding uncertainty, compared against quantized CLIP and DINO features. LangSplat\cite{14Qin_Li_Zhou_Wang_Pfister_2023} uses a scene-wise language autoencoder to distinguish clear object boundaries in feature images. Feature3DGS\cite{16Zhou_Chang_Jiang_Fan_Zhu_Xu_Chari_You_Wang_Kadambi_2023} proposes a parallel N-dimensional Gaussian rasterizer to distill high-dimensional features for tasks like editing and segmentation.
To ensure consistency across views, Gaussian Grouping\cite{17Ye_Danelljan_Yu_Ke_2023} integrates 2D mask predictions from models like SAM with 3D spatial constraints, enabling accurate object reconstruction and segmentation in open-world scenes.

Our approach aligns with these methods but leverages DINO\cite{Caron_Touvron_Misra_Jegou_Mairal_Bojanowski_Joulin_2021} for feature extraction, improving segmentation completeness and consistency in dynamic 3D environments.

\section{METHOD}
As illustrated in Figure \ref{fig:framework}, the framework consists of two main components: 3D Gaussian splatting for reconstruction and TOAO extraction. In the first part, our work builds upon LangSplat, with a key modification tailored to in-hand object reconstruction. Specifically, we propose a novel paradigm: a "static camera, moving object" setup. This paradigm requires specialized preprocessing of the captured images to ensure accurate object reconstruction during manipulation.

In the second part, we enhance the TOAO extraction process by employing a two-stage query mechanism using DINO. Initially, the entire object is queried to obtain a comprehensive understanding of its geometry. Subsequently, we perform a second query focused on identifying the most relevant local parts of the object. This dual-query strategy ensures a more accurate and task-specific TOAO extraction, leading to improved performance in downstream manipulation tasks.

\subsection{Preprocessing of In-hand Object Reconstruction}

In eye-in-hand systems, the spatial relationship between different coordinate frames is typically defined by the transformation matrix Equation \ref{eq}. The conventional paradigm for object reconstruction involves a "static object, moving camera" approach, ensuring consistent scene geometry throughout the reconstruction process. In this scenario, the transformation matrix ${}^{W}_{O}\mathbf{T}$, representing the object’s pose relative to the world frame, remains constant, as expressed by:

\begin{equation} {}^{W}_{O}\mathbf{T} = {}^{W}_{C}\mathbf{T} \ {}^{C}_{O}\mathbf{T} \label{eq} \end{equation}
where ${}^{W}_{C}\mathbf{T}$ describes the transformation from the camera frame to the world frame, while ${}^{C}_{O}\mathbf{T}$ represents the transformation from the object frame to the camera frame. This decomposition ensures that ${}^{W}_{O}\mathbf{T}$, the object's pose in the world frame, is easily calculated by knowing the relative transformations between the camera, object, and world. This method benefits from the stability provided by keeping the object stationary, simplifying the 3D scene reconstruction.

In contrast, our proposed paradigm adopts a "static camera, moving object" framework. Here, the camera remains stationary, keeping ${}^{W}_{C}\mathbf{T}$ constant, while the object dynamically moves within the scene. Consequently, the relative transformation ${}^{W}_{O}\mathbf{T}$ now varies over time, requiring real-time updates to track the object's changing pose in the world. This shift introduces new challenges in the reconstruction process, as we must continuously manage the changing transformations and adjust the reconstruction pipeline to accommodate the object's movement.

The key innovation in this paradigm is the dynamic separation of the object and end-effector from the scene while filtering out background components that are static relative to the world frame but move relative to the end-effector. As depicted in Figure \ref{fig:framework}, we utilize a combined depth and semantic segmentation approach to isolate the object. Specifically, we apply a mask that preserves the object’s features if the depth information satisfies certain conditions, ensuring that only valid data is retained.

The depth coverage ratio, $R_d$, is calculated by comparing the area of valid depth pixels, $A_d$, against the total area of the object mask, $A_m$, as shown in Eq. \ref{eq_depth_mask}:
\begin{equation}
    \begin{aligned}
        R_d & \ = \ \ \frac{A_d}{A_m} \label{eq_depth_mask} \\
        R_d \geq \theta_d & \implies \text{Mask is retained}
    \end{aligned}
\end{equation}
Here, $\theta_d$ represents the predefined threshold for depth coverage. If the ratio $R_d$ exceeds this threshold, the mask is retained for further processing, ensuring that the object's structure is accurately captured without interference from irrelevant background details.

This approach not only addresses the challenges posed by the moving object paradigm but also improves the reconstruction accuracy by dynamically adjusting the masking process based on depth information, making it more robust in real-world environments.

\subsection{Conditional 3D Object Extraction}

Our approach builds upon the 3D Gaussian Splatting (3DGS) method, which leverages Structure-from-Motion (SfM) techniques such as COLMAP to extract initial feature points. However, a critical limitation of 3DGS is its reliance on sparse feature points derived from the hand, which are not only fewer in number but often of lower quality compared to the dense feature points available across the entire scene. This sparsity introduces noise and artifacts into the final reconstruction, negatively affecting the performance of subsequent tasks. Moreover, 3DGS lacks effective spatial grouping mechanisms within the object itself, resulting in inconsistencies across different viewpoints.

These inconsistencies primarily arise from the fact that the visually identifiable regions of an object—particularly those relevant to task-oriented affordances—can vary significantly across views. As a result, the robot may focus on irrelevant parts of the object, which is problematic in real-world task scenarios. This misalignment occurs because 3DGS is inherently trained on localized image crops, limiting the contextual awareness of the CLIP embeddings regarding whether a particular crop belongs to the target object.

To overcome these limitations, GauTOAO introduces a language-guided approach to extract a complete 3D object mask. We integrate 3D DINO embeddings (self-distillation without labels), which have demonstrated strong object-awareness and the ability to effectively distinguish between foreground and background elements\cite{Caron_Touvron_Misra_Jegou_Mairal_Bojanowski_Joulin_2021,Amir_Gandelsman_Bagon_Dekel,Oquab_Darcet_Moutakanni}. By incorporating DINO embeddings into the 3DGS framework during inference, we improve object segmentation, enhancing both recognition and spatial coherence across views.

In the LangSplat framework, CLIP provides three levels of semantic features, each offering different degrees of granularity. Our process begins with the coarse-level features to obtain an initial object localization based on a language query. This ensures that no major parts of the object are overlooked, even in cases with low semantic distinction. We then refine this localization by generating a foreground mask, achieved by thresholding the first principal component of the top-down rendered DINO embeddings. By restricting the query to this mask, we ensure that the selected 3D points are contextually relevant to the task at hand.

Once the initial 3D point localization is established, we refine the object mask further. This is achieved by constructing an object-centric point cloud around the localized 3D point using multi-view deprojection of Gaussian-distributed points. The object mask is iteratively expanded by including neighboring points that fall within a defined threshold of DINO similarity, a process analogous to flood-fill algorithms used in image segmentation. The final output is a set of 3D points that accurately delineates the target object.

After extracting the full object, we employ higher-resolution semantic features from CLIP to perform more precise queries within the object itself. This allows us to pinpoint the exact region most relevant to the task, ensuring a more accurate and task-oriented 3D reconstruction.

\subsection{Task-Oriented Relative Pose Estimation}

Building upon the understanding of the in-hand object provided by GauAoTo, we establish a robust perceptual foundation for higher-level robotic decision-making and planning tasks. However, due to calibration errors in the transformation matrix ${}^{W}_{C}\mathbf{T}$ and perception errors in ${}^{C}_{O}\mathbf{T}$, direct usage of these matrices may not always provide the necessary precision for downstream tasks. To address this, we compute the relative position of the object’s target part with respect to the end-effector coordinate frame, ensuring more accurate manipulation during downstream operations.

The relationship between these transformations can be expressed as follows:

\begin{equation} {}^{W}_{O}\mathbf{T} = {}^{W}_{E}\mathbf{T} \ {}^{E}_{O}\mathbf{T} \label{eq} \end{equation}
where ${}^{W}_{O}\mathbf{T}$ represents the transformation from the object frame to the world frame, ${}^{W}_{E}\mathbf{T}$ represents the transformation from the end-effector frame to the world frame, and ${}^{E}_{O}\mathbf{T}$ is the transformation from the object frame to the end-effector frame. We define ${}^{E}_{O}\mathbf{T}$ as the task-oriented affordance transformation matrix, essential for robotic manipulation tasks.

This transformation matrix can be further parameterized as ${}^{E}_{O}\mathbf{T}(t, p)$, where $t$ refers to the specific task and the corresponding object, and $p$ denotes the pose of the object in the gripper. By utilizing the outputs from GauAoTo, augmented with knowledge from large models or expert systems, we can accurately compute ${}^{E}_{O}\mathbf{T}(t, p)$, which can then be used for precise task execution in downstream robotic operations.

\section{EXPERIMENT}
\subsection{Setup Details}
Our experimental setup consists of a UR3 robotic arm, which features six degrees of freedom, equipped with a DH Robotics gripper. For visual perception, we utilize an Intel RealSense D435i depth camera, which provides both RGB and depth information. The system is powered by a workstation equipped with an AMD Ryzen 7950X processor and an NVIDIA GeForce 3090Ti GPU.

\begin{figure*}[!ht]
    \centering
    \begin{subfigure}{0.21\textwidth}
        \includegraphics[height=4cm]{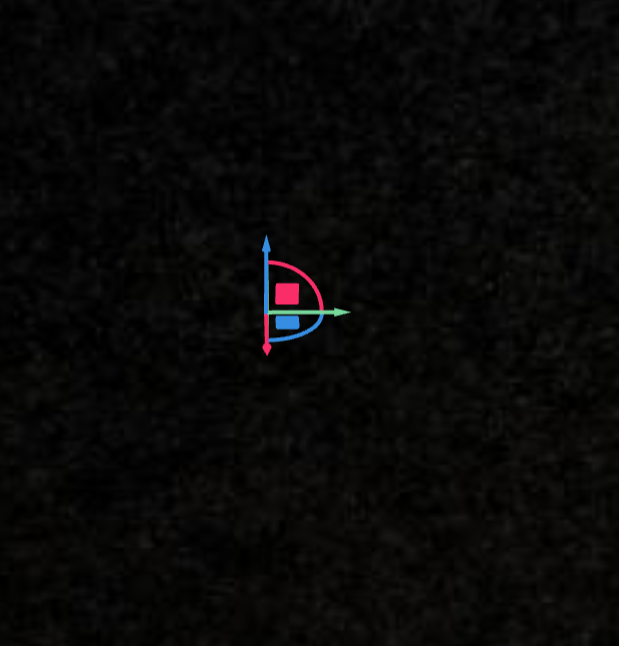}  
        \caption{}
        \label{41}
    \end{subfigure}
    \begin{subfigure}{0.23\textwidth}
        \includegraphics[height=4cm]{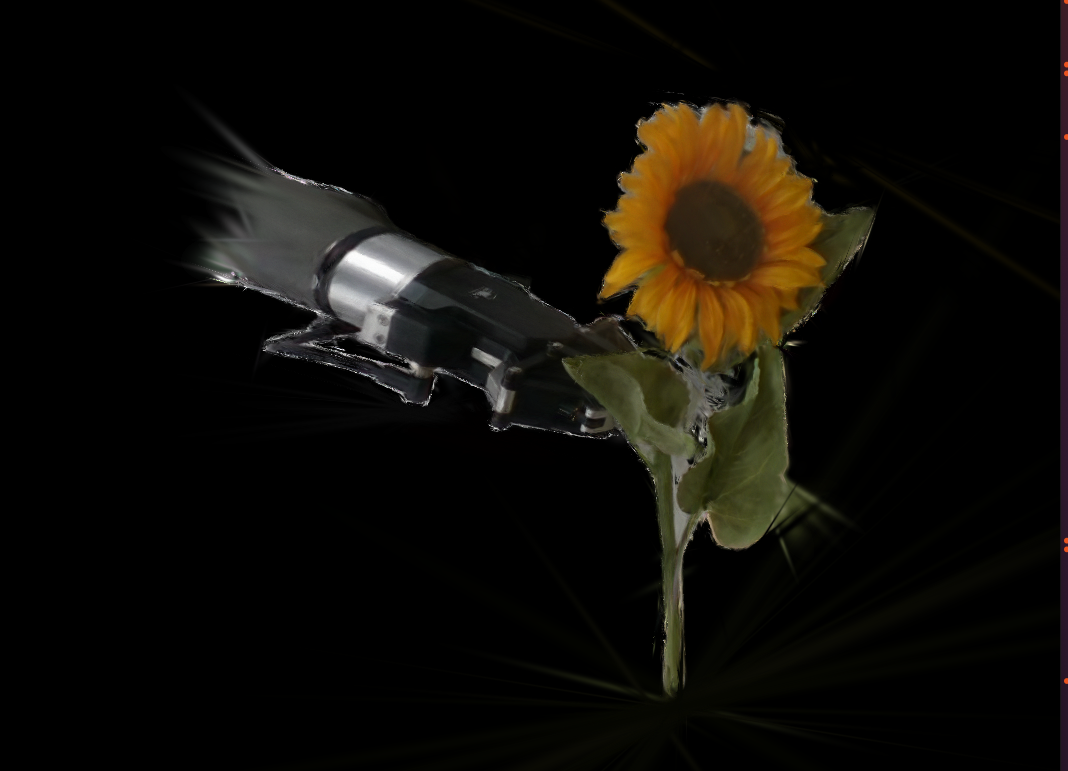}  
        \caption{}
        \label{42}
    \end{subfigure}
    \hspace{1.25cm}  
    \begin{subfigure}{0.15\textwidth}
        \includegraphics[height=4cm]{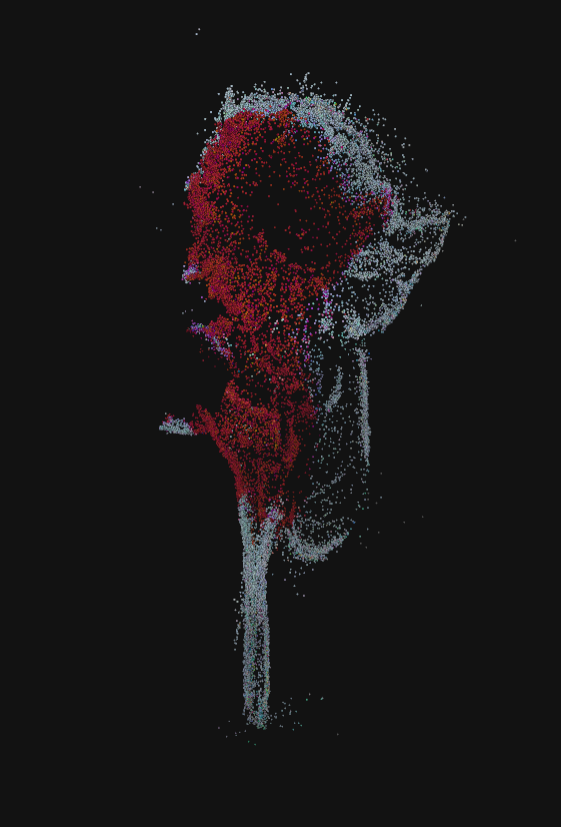}  
        \caption{}
        \label{43}
    \end{subfigure}
    \begin{subfigure}{0.15\textwidth}
        \includegraphics[height=4cm]{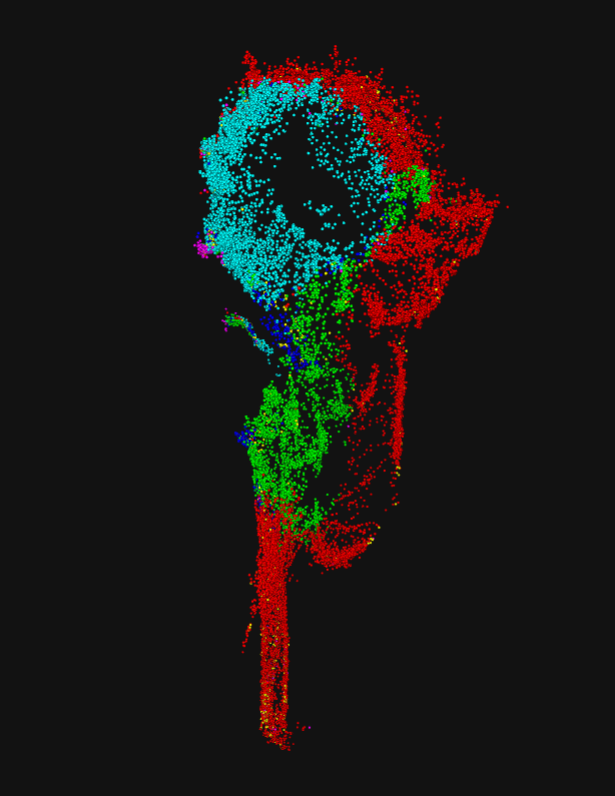}  
        \caption{}
        \label{44}
    \end{subfigure}

    \caption{Comparison of 3D reconstruction and TOAO extraction results for LERF, LangSplat, and GauTOAO.}
    \label{fig:reconstruction_comparison_1x4_fixed_height}
\end{figure*}

\subsection{LLM Interface for TOAO}

Given that the focus of our work is the extraction of Task-Oriented Affordance of Objects (TOAO), the upstream decision-making process, as well as prior knowledge, is derived from either expert input or large language models (LLMs). This structured approach enhances system stability and ensures consistency in affordance extraction. The following details our specific prompt instructions, designed to guide this decision-making process.

The prompt is crafted as follows:

\begin{quote}
\texttt{"Answer the question as if you are a robot with an object in your gripper. Follow the exact format. You will receive two pieces of input: 'O' representing the object in your gripper and 'T' representing the task you need to perform with the object. Provide a response, identifying a specific part of object 'O' that is most useful for completing task 'T'. Only provide the specific part of 'O'."}
\end{quote}

This prompt formulation ensures that the robot selects the most relevant part of the object based on task requirements, optimizing the task-oriented affordance extraction process. By leveraging LLMs or expert-derived knowledge, we ensure that the robot's decision-making aligns with real-world constraints, promoting both flexibility and reliability in dynamic environments.

The result is as follows:
\begin{quote}
\texttt{O: a bouquet of sunflowers\\
T: enjoy it\\
-A: The blossoms.\\}
\texttt{
O: a bouquet of sunflowers\\
T: put it into the vase\\
-A: The stem.\\}
\end{quote}

\subsection{Results of Relevancy }

We conducted a detailed experimental analysis on the "Sunflower Handling" task. The specific steps of the experiment were as follows: the robot's wrist joint performed rotational movements while capturing images (a total of 156 RGB-D images were collected in this experiment), followed by image preprocessing, semantic 3D reconstruction, and TOAO detection.

For comparison, we employed two mainstream baselines: LERF and LangSplat. The results of the semantic 3D reconstruction are shown in Figure \ref{fig:reconstruction_comparison_1x4_fixed_height}.

We also provide a quantitative comparison of semantic segmentation and localization accuracy between the three methods (LERF, LangSplat, and our proposed GauTOAO) in Table \ref{tab:realworld3}.


\begin{table}[!htb]
\centering
\caption{Quantitative comparisons of semantic segmentation and localization accuracy across LERF, LangSplat, and GauTOAO in our scenarios.}
{\label{tab:realworld3}
\begin{tabular}{ccc}
\toprule
Method &mIoU (\%) & Accuracy (\%) \\
\midrule[0.5pt]
LERF  & NAN     & NAN       \\
\midrule[0pt]
LangSplat & 52.7        & 58.2       \\
\midrule[0pt]
GauTOAO    & \textbf{61.4}          & \textbf{81.6}       \\ \bottomrule
\end{tabular}}
\end{table}

As shown in Figure \ref{41}, the LERF method struggled in the object reconstruction task. Due to the lack of sufficient feature points, it failed to capture enough camera views that meet the requirements of Structure-from-Motion (SfM), resulting in unsuccessful reconstructions and ineffective TOAO extraction.

LangSplat demonstrated improved reconstruction results, as shown in Figure \ref{42}. However, due to inconsistencies in certain object regions, its performance in TOAO extraction was suboptimal, as depicted in Figure \ref{43}.

Our method, GauTOAO, illustrated in Figure \ref{44}, significantly improved TOAO extraction accuracy. The two-stage detection process, incorporating DINO, allowed for more robust feature extraction, enhancing the accuracy of TOAO detection and ultimately improving the success rate in subsequent downstream manipulation tasks.

\section{CONCLUSIONS}

In this work, we shift the focus from the well-studied problem of Task-Oriented Grasping (TOG) to Task-Oriented Affordance of Objects (TOAO). While it is essential for robots to understand how to grasp objects, it is equally, if not more, important for them to comprehend how to use those objects in various tasks. We propose GauTOAO, a Gaussian-based framework for Task-Oriented Affordance of Objects, which enables robots to identify and attend to different parts of an object based on the task at hand. To the best of our knowledge, we are the first to perform semantic reconstruction in non-static coordinate frames, such as the end-effector of a robotic arm. Experimental results validate the effectiveness of our approach, demonstrating its ability to accurately identify task-relevant regions of objects in hand.

Furthermore, we believe that this work holds fundamental significance for the field of embodied intelligence. By enabling robots to truly understand the objects they are interacting with, we take a crucial step towards more intelligent and adaptable robotic systems.

\section{ACKNOWLEDGMENT}
The work is supported in part by the National Natural Science Foundation of China (No. 62176004, No. U1713217), Intelligent Robotics and Autonomous Vehicle Lab (RAV),
the Fundamental Research Funds for the Central Universities, and High-performance Computing Platform of Peking University.

\newpage
\bibliographystyle{ieeetr} 
\balance
\bibliography{refs} %

\end{document}